\documentclass[review]{elsarticle}

\usepackage{lineno,hyperref}
\usepackage{algpseudocode}
\usepackage[english]{babel}
\usepackage[utf8]{inputenc}
\usepackage{algorithm}
\usepackage{ifthen}

\usepackage{graphicx}
\usepackage{multirow}
\usepackage{subfloat}
\usepackage{float}

\usepackage{babel}
\usepackage{fancyhdr}
\usepackage{subcaption}
\usepackage{geometry}
\usepackage{makeidx}
\usepackage{amssymb}
\usepackage{eucal}

\begin{document}

\begin{frontmatter}

\title{A Simulation Model Demonstrating the Impact of Social Aspects on Social Internet of Things}

\author{Kashif Zia}
\address{Sohar University, Sohar, Oman}

\begin{abstract}
  In addition to seamless connectivity and smartness, the objects in the Internet of Things (IoT) are expected to have the social capabilities -- these objects are termed as ``social objects''. In this paper, an intuitive paradigm of social interactions between these objects are argued and modeled. The impact of social behavior on the interaction pattern of social objects is studied taking Peer-to-Peer (P2P) resource sharing as an example application. The model proposed in this paper studies the implications of competitive vs. cooperative social paradigm, while peers attempt to attain the shared resources / services. The simulation results divulge that the social capabilities of the peers impart a significant increase in the quality of interactions between social objects. Through an agent-based simulation study, it is proved that cooperative strategy is more efficient than competitive strategy. Moreover, cooperation with an underpinning on real-life networking structure and mobility does not negatively impact the efficiency of the system at all; rather it helps.
\end{abstract}


\end{frontmatter}

\section{Introduction} \label{sec:introduction}

\textit {Internet of Things (IoT)} \cite{gubbi2013internet} is claimed by many as the thing of the future \cite{park2016recent}. It is and would be more pervasive and socially embedded than any other recent technology. Already, our society has experienced undesirable outcomes of blindly adopting Internet-enabled mobile computing and social networking technologies \cite{valenzuela2009there}. Since the scale and embeddedness of objects of a potential IoT landscape are limitless in nature, a focus on human needs and social dimension has become a mandatory requirement. Already, the established notion of ``social objects'' (of Social Internet of Things (SIoT)) provides an opportunity to explore this dimension. Irrespective of the ethical discussions about technologies changing \textit{humanistic} values \cite{valenzuela2009there}, no one could overturn evolutionary progression seen in recent history. This is also valid for the \textit{Social} Internet of Things (SIoT). Since the scale of an IoT landscape is huge -- both horizontally and vertically --, so must be its impact. Therefore, envisaging the outfalls of such a landscape is extremely important, which is the focus of this paper. 

The relevance of social capabilities for the objects of IoT is evidenced in the growth of IoT itself, where the discussion on \textit{things} is now shifting from {\it smart} objects to the {\it acting} objects. ``Smart objects" are capable to communicate with the human social networks, while ``acting objects" are pseudo-objects able to represent human beings, sensing and acting on their behalf. There are already a number of applications in the above two domains. However, the real challenge is the futuristic ``social objects", having the capability to build their own social network. If not carefully modeled, objects interacting, acting and influencing their self-managed networks may turn out to be counter-productive or even harmful for the society \cite{atzori2014smart}. 

Setting aside the debate about its form-factor and usefulness, the transformation of current IoT towards futuristic network constituted by so-called ``social objects'' is inevitable \cite{atzori2015social}. According to Afzori et. al. \cite{atzori2014smart}, the challenges towards this quest can be grouped into three categories: (i) Designing conceptual frameworks and software platforms to enable socially interacting objects, (ii) introducing feasible mechanisms that cater for trust management between two interacting objects, and (iii) conceptualizing social paradigms to enable network navigability across trillions of potential objects to interact with.

The realization of challenge 1 is dependent on feasible solutions of second and third challenge. Challenge 2, as stated in \cite{atzori2014smart}, though only relates to one of the dimensions (i.e. trust) in which social interaction is dependent. However, it provides a clue about what possibly can be the other dimensions. It can be generalized that challenge 2 relates to \textit{social capabilities} (or incapabilities) directing the dynamics of interaction in a social setting. Whereas, challenge 3,  in fact, a generalized description of \textit{accessibility} of an object to another, is a relaxed notion of network access and connectivity. 


The exceptional growth of social networks in the last years has influenced many researchers. This is particularly true for social IoT, as evidenced from the above three challenges, which are tightly intermingled with social networking concepts. However, we propose that to argue about social IoT, we need to think beyond social networks. Motivated by the seminal concept introduced in \cite{atzori2014smart}, we take the liberty and postulate a guiding outline about the capabilities of a social object, which are: 
\begin {enumerate}
\item Autonomous nature of social objects takes decentralized decisions using the knowledge available at the local level.
\item Social objects broadcast their service and make its presence feel. 
\item Social objects trust their peers and inter-object interaction is supported.  
\end {enumerate} 
The above three features act as an underpinning of our quest towards analyzing the impact of social aspects for SIoT. The social objects must be autonomous and self-organized. Having an underpinning on the concept of agency \cite{helbing2012agent}, the social objects must possess a ``self-organizing collective behavior not resulting from the existence of a ``central controller'', but due to their own interactions with the other agents.'' \cite {caram2015cooperative}. The interactions require awareness of the proximity. Hence, the first two features are indigenous to agent-based modeling paradigm, wherever it is used to model IoT (as that of this paper). As of \cite{atzori2014smart}, Peer-to-Peer (P2P) computing \cite{loo2007peer} is taken into consideration as a typical application assuming that interacting peers have absolute trust on each other.

The services provided and requested by the peers are disparate, so is the network configuration. The configurations supported by the network are of different types. By default, the peers can communicate with their peers in the neighborhood. However, a ``small-world network'' configuration \cite{watts1998collective}, in which a few long-distance connections augment an otherwise ``regular network'', can mimic a realistic social possibility, in which the services available from distant peers are supported through a few long distant peers. It is assumed that long distant communication requires more resources. Therefore, only a few randomly chosen peers (dependent on the beta value of small-world) are considered as peers. Also, these peers would only be invoked if there is no possible service provider in the locality. A further enhancement only takes ``legitimate'' peers (friends) for the provisioning of service time. 


In this paper, the impact of social behavior on the interaction pattern of social objects is studied taking P2P resource sharing as an example application. The model proposed in the paper studies the implications of competitive vs. cooperative social paradigm while peers try to acquire shared resources/services. These contradictory behaviors are modeled through an agent-based model (ABM). In addition, to the proposed model adhering to the notion of ``social objects'' \cite{atzori2014smart}, our model supports asymmetric peers in terms of their capabilities and services configurations, and realization of a more realistic real-world networking and objects' mobility.

In the rest of this paper, Section \ref{sec:rw} presents the related work and motivation, section \ref {sec:model} presents the proposed model, followed by simulation and results in section \ref {sec:sim}, and section \ref {sec:conc} concludes the paper.

\section{Related Work} \label{sec:rw}

Computer networks offer amazing possibilities to exchange data and dealing with files \cite{gebali2015analysis}. Old communication systems sited resources on a centrally-managed server, which can be accessed by the client machines that connect to the central server (a "client-server" relationship). In contrast, P2P technology allows client machines on a network to share their own resources (file store, processing power, and peripherals.) with other network-connected machines with or without involvement of any centralize server \cite{popovskyy2017analysis}. Thus, it enables machines to act as a client as well as server at the same time. This influential technology makes it possible for the contents to be distributed widely without the requirement of the central facility of large resources in terms of computing power, storage or in particular, network bandwidth \cite{navimipour2015comprehensive}.

P2P technology enabled a revolution in machine-to-machine communication. While the machines are increasing exponentially, there is a desire to reduce their cost almost to nothing to enable communication between even very small and ordinary things around us. This endeavor has resulted in technologies and systems like Wireless Sensor Network, Internet of Things (IoT), Cyber-Physical Systems and Human-Agent Collectives (HAC) (sociotechnical systems) \cite{pticek2016beyond}. IoT has gained a tremendous interest not only from the researchers but also from the industry. According to CISCO \cite{evans2011internet}, the number of connected devices will surpass 50 billion by 2020, which indicates a huge market \cite{ihs}. Interest groups have been formed to define frameworks and standards for the IoT. Major IT companies have introduced a number of products and services based on IoT and invested substantial amount such as Nest \cite{Nest} by Google and SmartThings by Samsung \cite{smartthings}. A number of leading ICT organizations have introduced IoT solutions such as Amazon Web Services, Ericsson, Huawei, IBM, and others. Huawei's Ocean Connect \cite{huawei} introduced a sophisticated IoT platform recently.

The latest application of such technologies exists in Industry 4.0, Industry 4.0 \cite{bartodziej2017concept} is a European initiative promising to transform industrial system of the future by integrating new technologies towards sustainability, efficiency and safety. According to \cite{camarinha2017collaborative}, collaboration issues are one of the most demanding aspect of this movement. Many social aspects are identified such as strategic decision-making, behavioral and trust modeling, collaborated group achievement and optimization, and evolving network dynamics. Without any doubt, IoT as the underlying network of Industry 4.0 needs to be \textit{at least} socially-aware thus acting as an enabler to address these challenges. 

Authors in \cite{cervantes2016could, worthy2016trust} investigated the impact of IoT technologies, and the application on human values. In particular, the importance of trust in technology is highlighted to improve person-to-person communication using IoT medium. Another model evaluating honesty is presented in \cite {jayasinghe2017computational}. The overall context of users owning IoT devices is implemented as a collaboration module in \cite{kum2015design}. An IoT based study to verify users based on the pattern of their activities is presented in \cite{anjomshoa2017social}, thus emphasizing the importance of social aspects.

The study of the overall working mechanism of a social system {\it being competitive, cooperative, or mixed-type} has been a topic of interest \cite {hauert2006cooperation}. Cooperation with peers is found in many natural systems \cite{dorigo2006ant}. Human societies, trained to be competitive \cite {fehr1999theory} are learning to appreciate cooperation as the winning strategy \cite{barreira2013evolution}. Hence, in many research contributions, the conditions of transforming a population from competitive to cooperative mode are studied. 

For example, the influence of activity like migration for the outbreak of cooperation is studied in \cite{schweitzer2012optimal}. A simplistic interaction situation of iterated Prisoner's dilemma is implemented. However, the model is population-based and does not focus on individual characteristics of interacting entities which are necessary to investigate the emergence of cooperation. The specifics of formation of reciprocal appreciation in small groups in studied using an agent-based model \cite{koponen2016formation}. Although the intuition of the model is based on social interaction idea, it does not provide a functional specification of activity performed through inter-agents interaction. 

Authors in \cite {caram2015cooperative} presented a model of interaction in a P2P network using a multi-agent-based system. Authors has presented a comparison between interaction efficiency whereby the model of interaction is influenced by the social dimensions of competitiveness vs. cooperation. According to the model, while sharing a common resource, the probability of interaction increases with increase in the difference between sizes of agents. The size of an agent is directly proportional to the portion of common resource acquired by an agent. Hence, a peer will cooperate with another peer if one can offer and the other can receive. However, the model is proposed for the symmetric setting in terms of agents' function, services, and network configurations. Motivated from the above model, in this paper, an agent-based model influenced by principles of social interactions in an asymmetric setting is proposed. It deviates from the above model in following way. Our model considers time instead of resources as decision making parameter. The motivation behind this change is that peers representing social objects would share the information about time spent in a state more naturally when compared with sharing the status of the common resource, due to an inherent conflict of interest.

\section {Models} \label {sec:model}

The purpose of the model to analyze the impact of social-like capabilities on the interaction pattern of social objects in P2P resource sharing scenario. A comparative study of competitive vs. cooperative behavior of peers is demonstrated with the help of an agent-based modeling method and simulation. The model adheres to the specifications of social objects. Each peer/agent acts as a decision making entity using only the knowledge it possesses without the assistance of a central coordinator. All agents acquire a uniform mobility model; the options are stationary, random-walk and profile-based mobility. Stationary and random-walk mobility is obvious. In profile-based mobility, agents build some random location in their surroundings to move to, and they move from one location to another. All agents are available and their operations/statuses are transparent to others. For simplicity, considering that that all the agents in a specific range (neighborhood) are equally accessible and trust each other unconditionally.

The services provided and requested by the peers are disparate, so is the network configuration. By default, the peers can communicate in their neighborhood (the regular network). However, a small-world network configuration \cite{watts1998collective}, in which a few long-distance connections augment an otherwise regular network, hence, able to mimic a realistic social network. It is assumed that long distant communication requires more resources. Therefore, only a few randomly chosen peers (dependent on the beta value of small-world) are considered as peers. Also, these peers would only be invoked if there is no possible service provider in the locality. A mesh network is also considered as the base case. Further, only the \textit{legitimate} peers (friends) are for the provisioning of service time, thus, imparting social-like capabilities. 

The models are closely entangled with the services' specifications, and agents' basic interactions capabilities. A comprehensive explanation of these specifications is given in Section \ref{sec:specs}. The agent-based models of competition and cooperation follow in Section \ref{sec:comp} and \ref{sec:coop}. A model of friendship with increasing repetitive interactions is presented in Section \ref {sec:friendship}.

\subsection {Peer Specifications} \label{sec:specs}

It is assumed that four services are provided by the network. Each service requires a time effort to complete it, termed as Service Completion Units (SCU), where a unit is equal to one iteration of the simulation. That is: $Serv1$ requires 25 SCU, $Serv2$ requires 50 SCU, $Serv3$ requires 75 SCU, and $Serv4$ requires 100 SCU. The fifth service, $Serv0$ is a different service, requiring no time (0 SCU), indicating that the agent at that time is serving another agent. A peer has a pre-defined duration of remaining off / on the network. It also has a pre-defined value \textit{serv0perc}, that represents for how long a peer would be serving another peer (if requested, able to provide the requested service and idle at that time) from its idle time in percentage.

A detailed description of all specifications is available in \cite {zia2018modeling}. Based on these specifications, a peer/agent would either perform its action in competitive or cooperative mode. There are six states of an agent on which the model of competitive and cooperative behavior are based, given in the following: 

\begin{enumerate}
\item {\it status 0}: when an agent is not on the network and it is physically in ``off'' state.
\item {\it status 1}: when an agent is in ``idle'' state representing no service required or delivered. 
\item {\it status 2}: when an agent in ``assign'' state, that is ready to go from an idle to an active state. 
\item {\it status 3}: when an agent is in ``search'' state, that is searching for {\it possible-service-providers} for the execution of the required service. 
\item {\it status 4}: when an agent is in ``request'' state, that is requesting for resources from a searched peer. 
\item {\it status 5}: when an agent is in ``proceed'' state, that is having completed the current service and is ready to execute another process cycle. 
\item {\it status 6}: when an agent would be in ``serve'' state, that is just serving another agent for a time equal to {\it service0perc} times the original {\it idle-time}. 
\end{enumerate}

The {\it duration in current status (DICS)} would track for how long a peer is in a specified status and helps in transiting from one state to another if certain conditions are met. Whenever a state changes, DICS = 0. Alternatively, it is incremented by 1.

\begin{figure*}
\centering
\includegraphics[width=.99\textwidth]{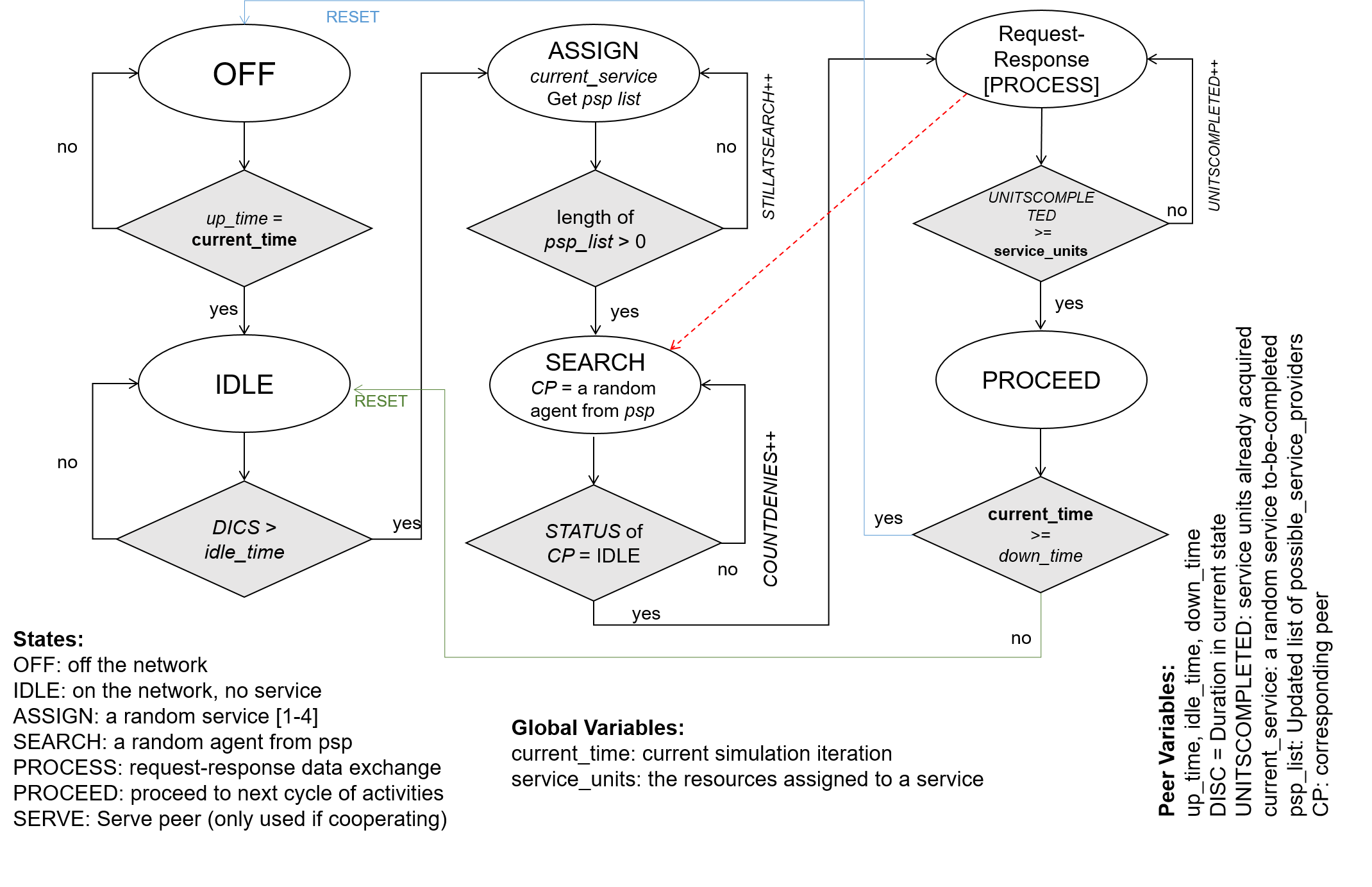}
\caption{Agent-Based Model of Peers in Competitive Mode. current\_time is iteration (a minute on a day). current\_service is one of the four possible services. psp contains all other agents in the neighborhood [strategy ``mesh'' (whole space) or ``regular'' (only neighborhood) or ``small-world'' (in a proximity defined by neighborhood and beta value)] who have completed the current\_service (evidenced by services\_completed table).}
\label{fig:pcompmode}
\end{figure*}

\begin{figure*}
\centering
\includegraphics[width=.99\textwidth]{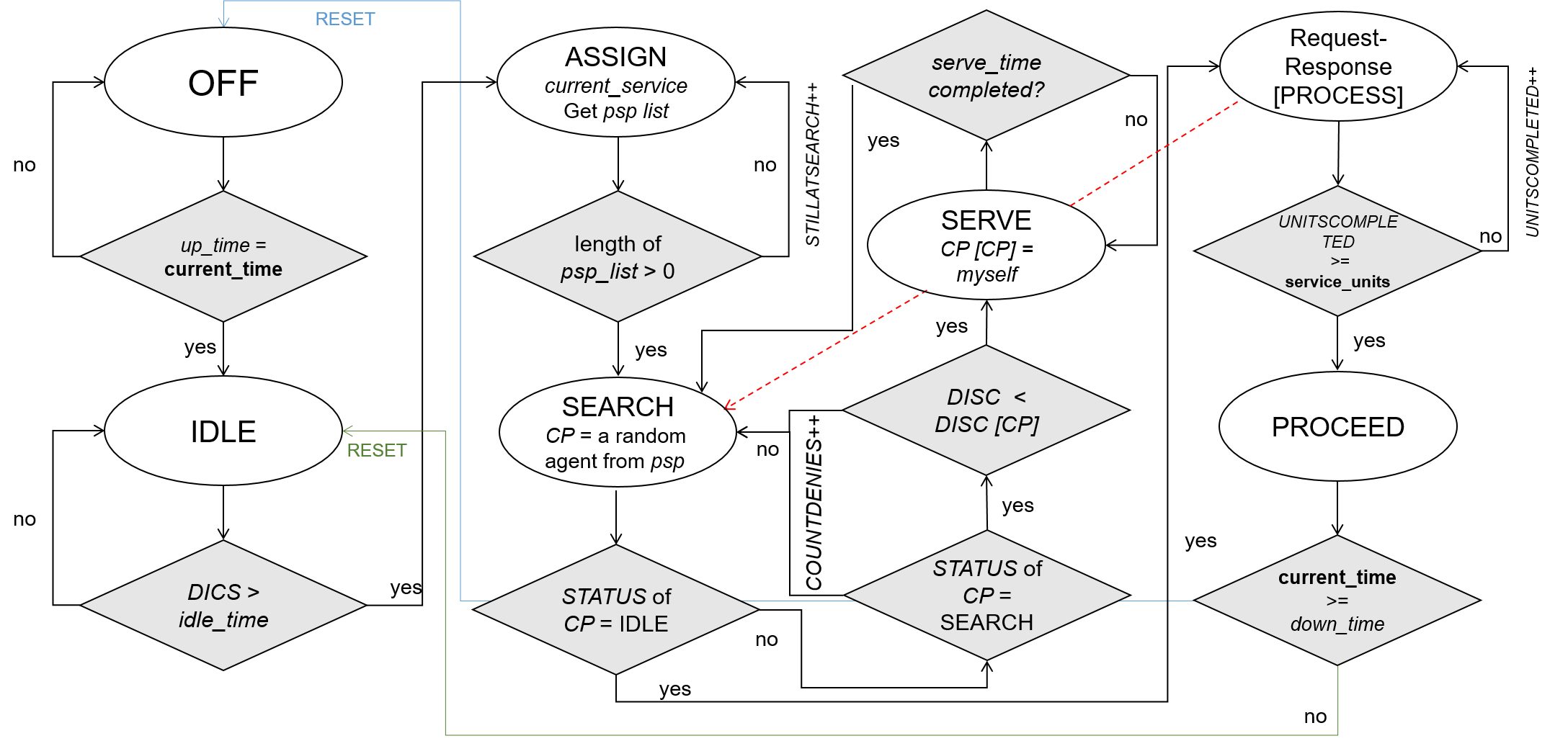}
\caption{Agent-Based Model of Peers in Cooperative Mode.}
\label{fig:pcoopmode}
\end{figure*}

\subsection {Model of Peers in Competitive Mode} \label{sec:comp}

Figure \ref {fig:pcompmode} represents the model of peers in competitive mode, which is detailed as:

\begin {itemize}
\item From being in the ``off'' state, an agent would transit into the ``idle'' state, if the current time (CT) of simulation is equal to {\it up-time} (the time at which a peer becomes part of the network every day) of the agent. 

\item  If a peer is in the ``idle'' state, it would transit to ``assign'' state, if the duration it had to be in the idle state is complete. 

\item In the ``assign'' state, a random service is assigned to the agent and it transits to "search" state. 

\item In the ``search'' state, an agent searches for {\it possible-service-providers (PSP)}, both in the neighborhood as well as a distance (depending on the network type used). These are agents who have already completed the desired service recently (after last status 0 to status 1 transition). If the search is successful, the agent transits to ``request'' state. 

\item In the ``request'' state, the chosen {\it corresponding peer} would reply against the request if its state is idle and the value of {\it consistency} (represents how consistent the peer is in terms of continuity of the connectivity) allows it. If such a reply is received, it is considered that one SCU of the current requested service has been completed. Otherwise, the agent transits back to ``search'' state. If {\it units-completed} is greater than or equal to SCU of the current service, the service is completed. The agent would record the service competed in {\it recent-services-completed} structure and would transit to ``proceed'' state. 

\item In the ``proceed '' state, a peer either transits back to state ``off'' or state ``idle'' depending on its {\it down-time} (the time at which a peer becomes unavailable on the network every day). All service-related parameters are also reset.
\end {itemize}

In the all above conditions, otherwise, the {\it duration-in-current-status} would be incremented.  

\subsection {Agent-Based Model of Peers in Cooperative Mode} \label{sec:coop}

When compared with competitive mode, the major refinement occurs in status 3, and a new state is introduced namely ``serve'' state with status = 6. Figure \ref {fig:pcoopmode} shows the model of peers in cooperative mode. Cooperation is expected to improve resource sharing efficiency of the system. In the following, we detail the differences in cooperative mode when compared to competitive mode.   

\begin {itemize}
\item In the "search" state, an agent searches for {\it possible-service-providers}, both in the neighborhood as well as a distance (depending on the network type used). If there are some possible service providers, and one of them is in status 1 or 6, the agent can readily start utilizing the service by entering into status 4.  If there is no service provider available, the peer will keep searching. In the case of two or more peers searching for each other, the conflict resolution mechanism would be invoked. The conflict would be resolved in favor of one of the interacting peers based on DICS. If DICS of the requester is greater than DICS of the respondent, the status of requester would be set to 6, which is, the requester would transit into {\it serve} state. The respondent status would be set to 4, thus, ready to utilize the service provider by the requester. An entirely opposite procedure would be followed if the DICS condition is false.   

\item In the ``serve'' state, a peer would just serve the other peer being idle for {\it x} iterations, where {\it x} is equal to {\it service0perc} multiplied with the original {\it idle-time}. 
\end {itemize}

\subsection {Model of Friendship (Restricted Cooperation)} \label{sec:friendship}

Let's name the agent as \textbf{A} being in memory, and $myneighbors$ is a list of current neighbors. Since agents may be not stationary, it is necessary to always search for the fresh neighborhood. To do so, we term agentset $neighbors$ as all the agents in prescribed radius excluding \textbf{A}, radius being one of model parameters. Every agent \textbf {a} in $neighbors$ set is processed. If \textbf {a} already exists in table $myneighbors$, we increment $repeated\_encounters$ value by 1 against its ID. If \textbf {a} is encountered for the first time, we insert its ID in table $myneighbors$ with $repeated\_encounters$ value set to 1.   

After a specific number of iterations (parameter consolidate\_frequency), we update $mycontacts$ table which has a subset of agents from $myneighbors$. The length of $mycontacts$ table have an upper bound value defined by parameter \textbf {k}. The number of agents to be added depends on the length of $myneighbors$ table times \textbf {k}. For example, if an agent has six neighbors, and $k = 0.5$, then the number of agents to be added in the $mycontacts$ table is three. A random agent, if it already does not exist in $mycontacts$, would be added from $myneighbors$ registering the iteration on which it is added against its ID. It is noteworthy that as the simulation progresses, the length of the $myneighbors$ table increases so does the length of the $mycontacts$ table. Lastly, the length of the $myfriends$ table has an upper bound value defined by parameter \textbf {m}. The number of agents to be added depends on the length of $mycontacts$ table times \textbf {m}. A random agent from $mycontacts$, if it already does not exist in $myfriends$ is added as a friend, registering the iteration on which it is added against its ID. The length of the $myfriends$ table would also increase with the $mycontacts$ table as the simulation progresses. Hence, the list of friends limits potential service providers, expected to decrease the resource sharing efficiency. 

\subsection {Mobility Models} \label {sec:mobility}

There are three \textbf{mobility} modes under which this mechanism operates.

\begin {enumerate}
\item Mobility 1: No mobility, i.e, all agents are stationary.
\item Mobility 2: Random walk; agents choose a direction to move randomly at each iteration. 
\item Mobility 3: Profile-based walk; agents build some random locations to move to, and they move from one location to another. 
\end {enumerate}

\section {Simulation and Results} \label {sec:sim}

\begin{figure}
\centering
\includegraphics[width=.49\textwidth]{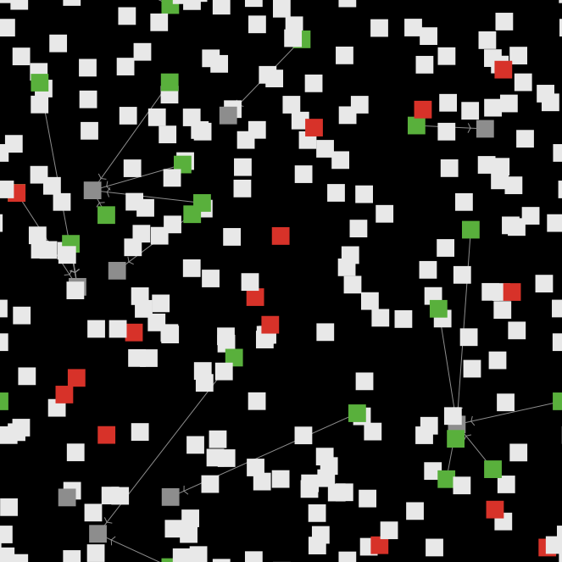}
\caption{Status of peers in COOPERATIVE MODE at iteration 14400 in case of 250 agents (radius = 5) using a small-world network (of beta = 0.2) and profile-based mobility. Agents in light gray are in offline mode (status 0). Agents in dark gray are in idle mode (status 1). Agents in red are in search mode (status 3). Status 2 is a transit mode where a service is assigned. Agents in green are in request mode (status 4). Status 5 and 6 are transit states again. The arrow originates from the requester and ends at a receiver.}
\label{fig:simspace}
\end{figure}

\subsection {Simulation Setup}

Simulation is implemented in NetLogo, an open-source agent-based modeling simulator \cite{wilensky2015introduction}. A series of parametric settings are possible (see Table \ref{tbl:setup}). Each case is shown in Table \ref{tbl:setup} is examined for three possible mobility modes, that is, stationary, random walk and profile-based walk. We start with the agent population of 250 agents (with a constant radius equal to 5) for a comparison between competitive and cooperative behavior. There are four possible services (shared resources). A service is either denied or completed in each iteration. Obviously, how efficiently a service is progressing is dependent on the ratio of denials and completions. Leaving a complete analysis of related results for future, here, we have focused on a single variable, that is, the number of ``nor served" requests, representing the mismatch between the requests generated and the servings which are activated due to limitations imposed by network and mobility model. Each case is run 100 times; each time with different random configurations. Results of 100 runs are averaged to present a normalized picture. 

\begin{table}[]
\scriptsize
\centering
\caption{Simulation Setup: Cooperative-R represents restricted cooperation}
\label{tbl:setup}
\begin{tabular}{|l| |l| |l| |l|}
\hline
{\bf Case} & {\bf Scenario} & {\bf Population} & {\bf Network Type}\\
\hline
Case 1 & Competitive & 100 & Mesh\\
Case 2 & Competitive & 100 & Regular\\
Case 3 & Competitive & 100 & small world (beta = 0.1)\\
Case 4 & Competitive & 100 & small world (beta = 0.2)\\
Case 5 & Competitive & 250 & Mesh\\
Case 6 & Competitive & 250 & Regular\\
Case 7 & Competitive & 250 & small world (beta = 0.1)\\
Case 8 & Competitive & 250 & small world (beta = 0.2)\\
Case 9 & Competitive & 500 & Mesh\\
Case 10 & Competitive & 500 & Regular\\
Case 11 & Competitive & 500 & small world (beta = 0.1)\\
Case 12 & Competitive & 500 & small world (beta = 0.2)\\
\hline
Case 13 & Cooperative & 100 & Mesh\\
Case 14 & Cooperative & 100 & Regular\\
Case 15 & Cooperative & 100 & small world (beta = 0.1)\\
Case 16 & Cooperative & 100 & small world (beta = 0.2)\\
Case 17 & Cooperative & 250 & Mesh\\
Case 18 & Cooperative & 250 & Regular\\
Case 19 & Cooperative & 250 & small world (beta = 0.1)\\
Case 20 & Cooperative & 250 & small world (beta = 0.2)\\
Case 21 & Cooperative & 500 & Mesh\\
Case 22 & Cooperative & 500 & Regular\\
Case 23 & Cooperative & 500 & small world (beta = 0.1)\\
Case 24 & Cooperative & 500 & small world (beta = 0.2)\\
\hline
Case 25 & Cooperative-R & 100 & Mesh\\
Case 26 & Cooperative-R & 100 & Regular\\
Case 27 & Cooperative-R & 100 & small world (beta = 0.1)\\
Case 28 & Cooperative-R & 100 & small world (beta = 0.2)\\
Case 29 & Cooperative-R & 250 & Mesh\\
Case 30 & Cooperative-R & 250 & Regular\\
Case 31 & Cooperative-R & 250 & small world (beta = 0.1)\\
Case 32 & Cooperative-R & 250 & small world (beta = 0.2)\\
Case 33 & Cooperative-R & 500 & Mesh\\
Case 34 & Cooperative-R & 500 & Regular\\
Case 35 & Cooperative-R & 500 & small world (beta = 0.1)\\
Case 36 & Cooperative-R & 500 & small world (beta = 0.2)\\
\hline
\end{tabular}
\end{table}

\subsection {Simulation Results}

\begin{figure}[htb]
\centering

\begin{subfigure}[b]{0.3\textwidth}
\includegraphics[width=\linewidth]{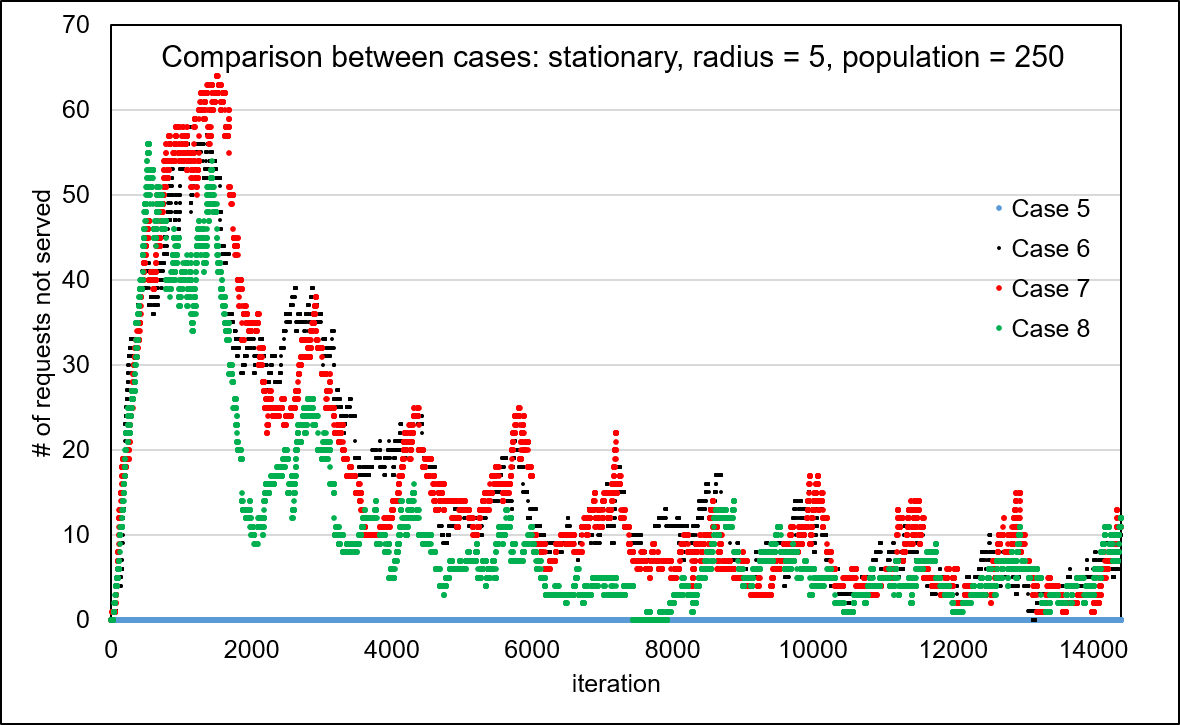}
\label{fig:compa}
\end{subfigure}
\hfill
\begin{subfigure}[b]{0.3\textwidth}
\includegraphics[width=\linewidth]{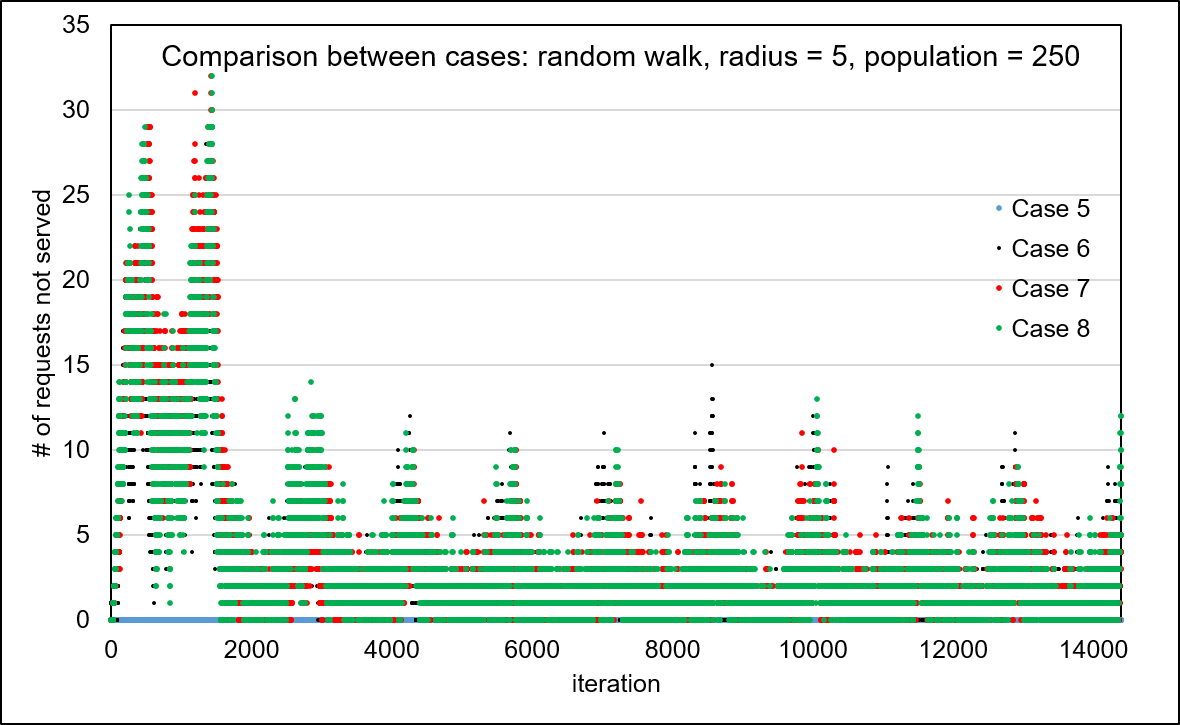}
  \label{fig:compb}  
\end{subfigure}
\hfill
\begin{subfigure}[b]{0.3\textwidth}
\includegraphics[width=\linewidth]{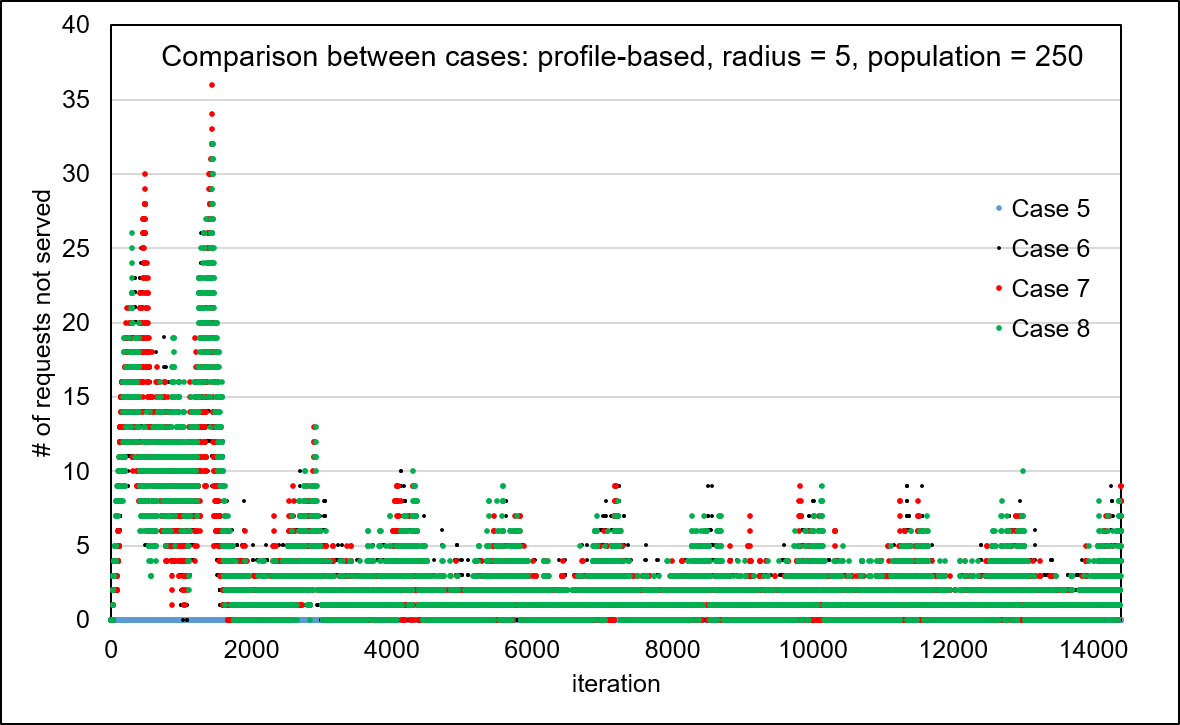}
\label{fig:compc}
\end{subfigure}

\caption{Simulation results: Competitive Mode (250 agents)}
\label{fig:fullcomp}
\end{figure}

\begin{figure}
\centering

\begin{subfigure}[b]{0.3\textwidth}
  \includegraphics[width=\linewidth]{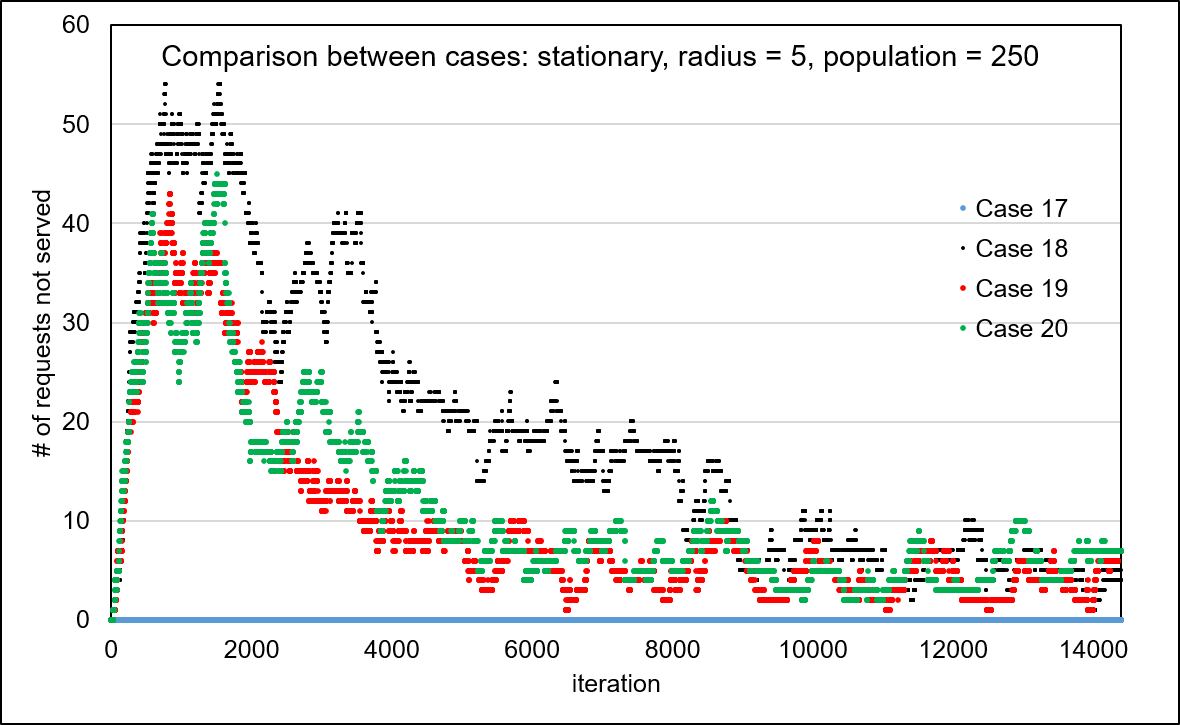}%
  \label{fig:coopa}%
   \end{subfigure}
\hfill
\begin{subfigure}[b]{0.3\textwidth}
  \includegraphics[width=\linewidth]{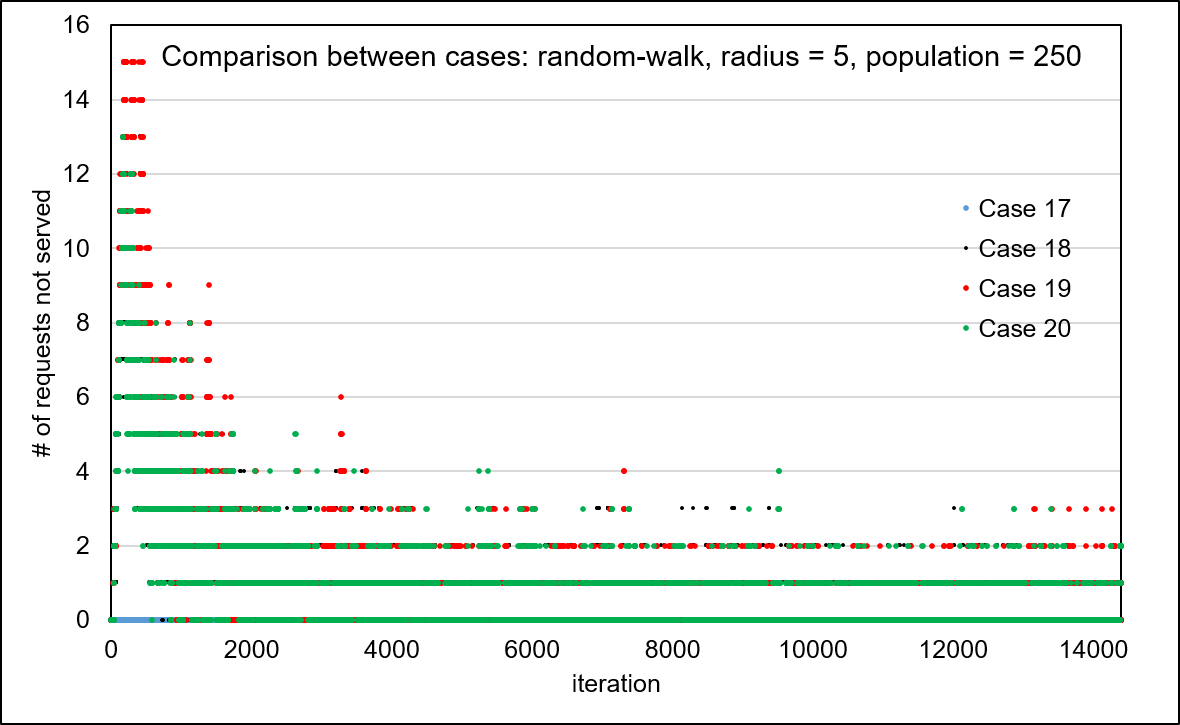}%
  \label{fig:coopb}%
 \end{subfigure}
\hfill
\begin{subfigure}[b]{0.3\textwidth}
  \includegraphics[width=\linewidth]{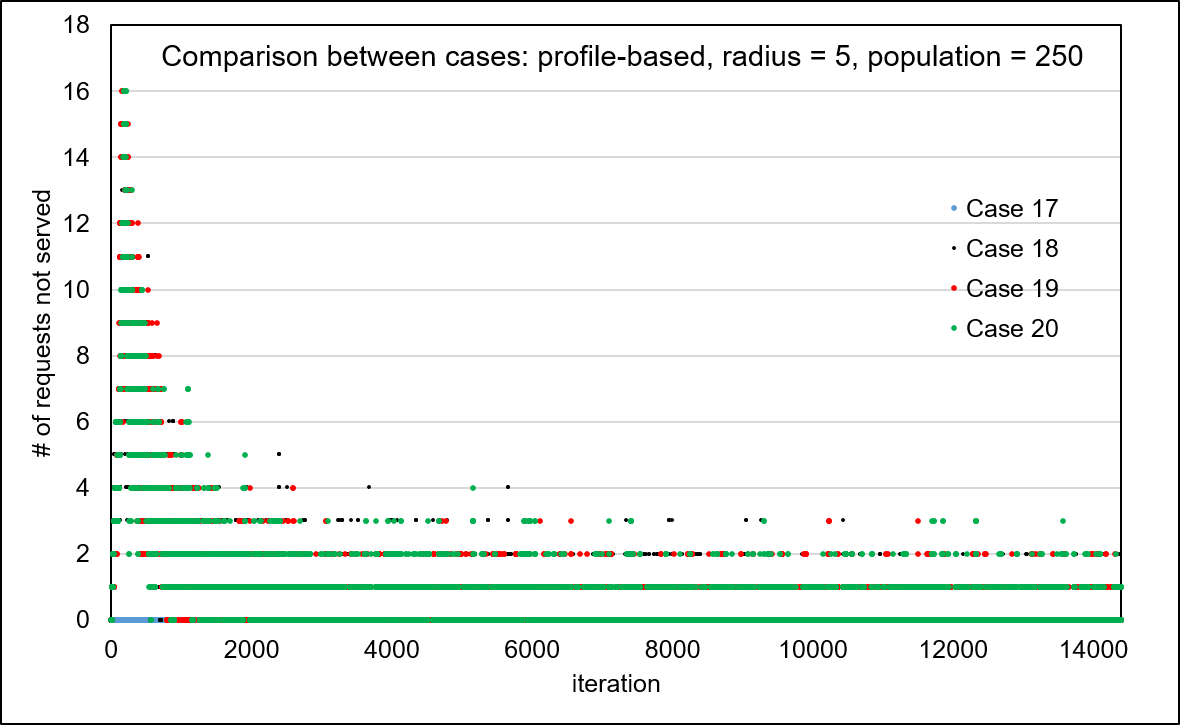}%
  \label{fig:coopc}%
 \end{subfigure}
 
\caption{Simulation results: Cooperative Mode (250 agents)}
\label{fig:fullcoop}
\end{figure}

\begin{figure}
\centering

\begin{subfigure}[b]{0.3\textwidth}
  \includegraphics[width=\linewidth]{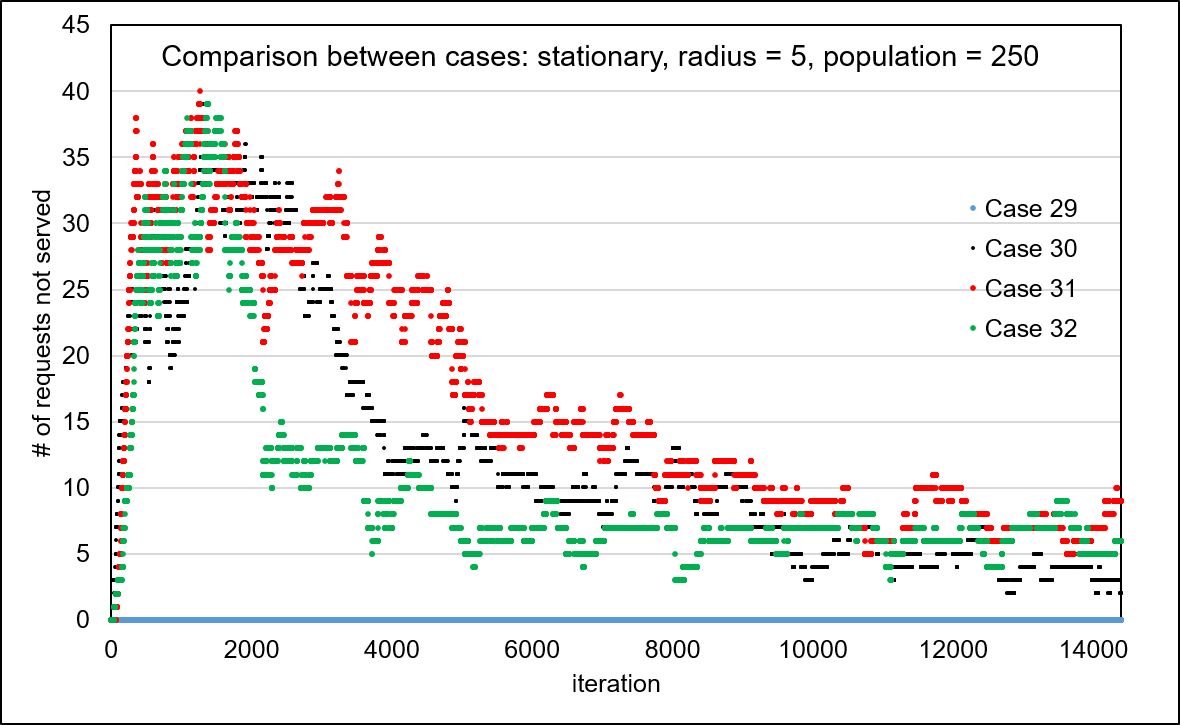}%
  \label{fig:coopra}%
 \end{subfigure}
\hfill
\begin{subfigure}[b]{0.3\textwidth}  
\includegraphics[width=\linewidth]{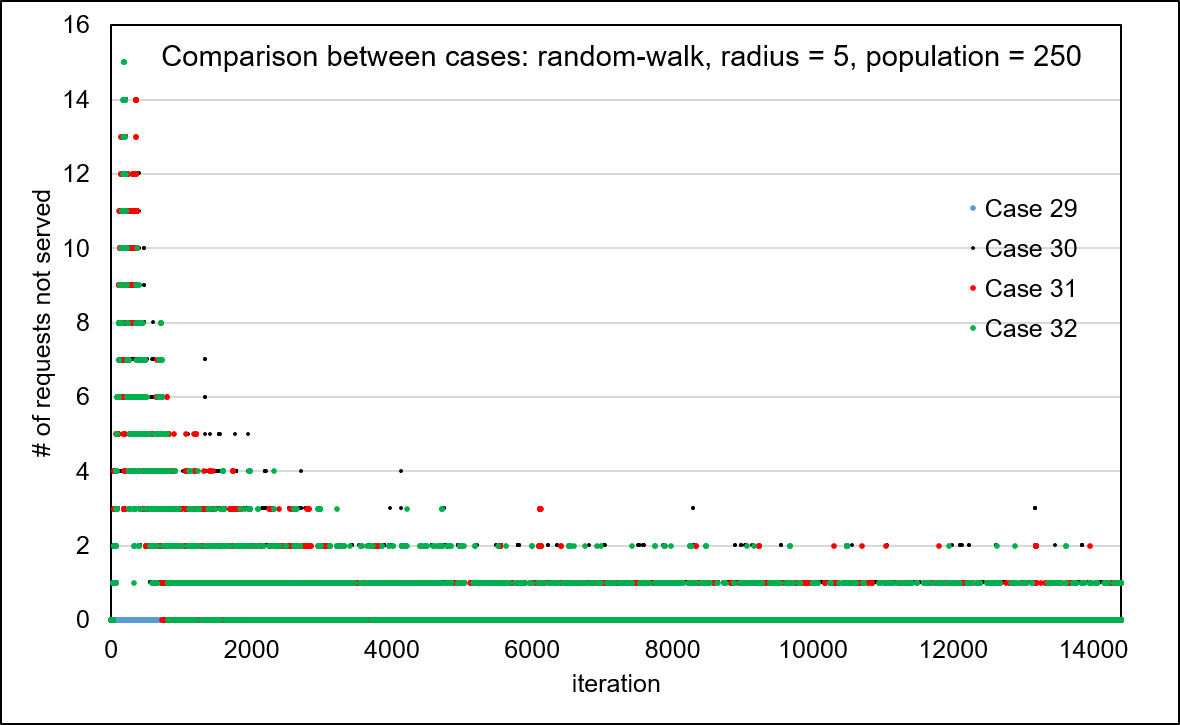}%
  \label{fig:coorpb}%
 \end{subfigure}
 \hfill
\begin{subfigure}[b]{0.3\textwidth}
  \includegraphics[width=\linewidth]{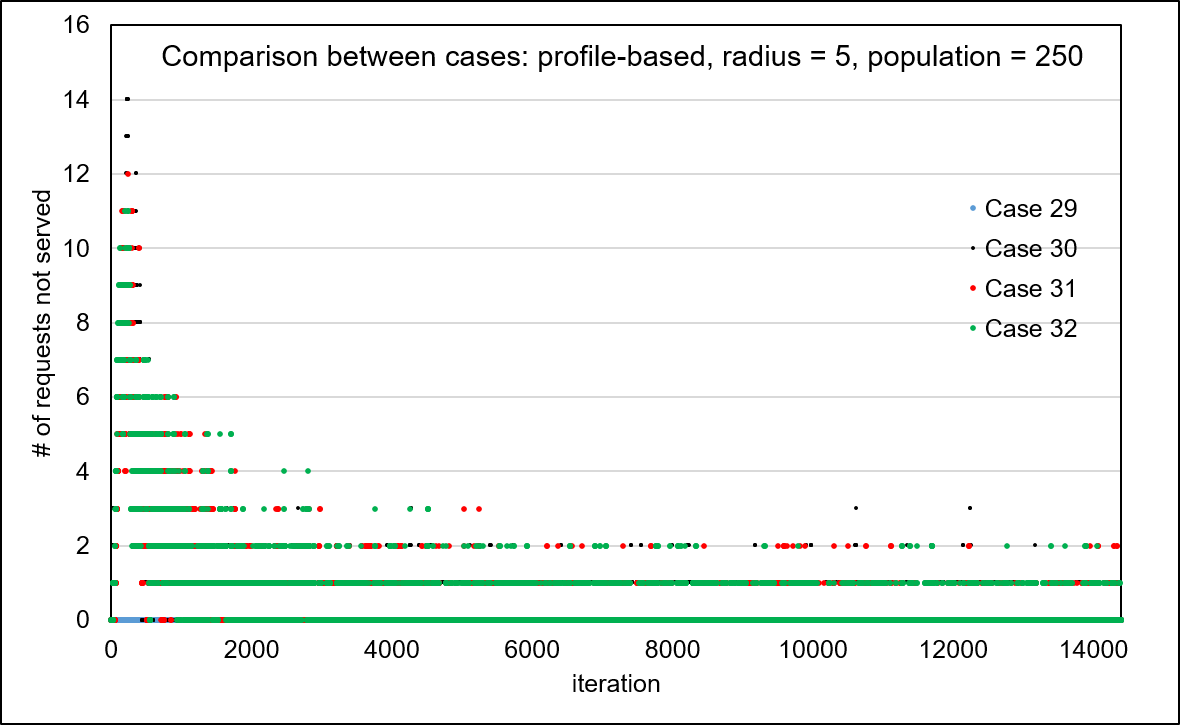}%
  \label{fig:cooprc}%
 \end{subfigure}

\caption{Simulation results: Cooperative Mode (Restricted) (250 agents)}
\label{fig:fullcoopr}
\end{figure}

For 250 agents, Figure \ref{fig:fullcomp} shows the simulation results when the strategy used is competitive; Figure \ref{fig:fullcoop} shows the simulation results when the strategy used is the cooperative; and Figure \ref{fig:fullcoopr} shows the simulation results when the strategy used is restricted cooperation. Irrespective of the strategy, mesh network is the best, and the number of request ``nor served" is always zero as indicated in case 5 (Figure \ref{fig:fullcomp}), case 17 (Figure \ref{fig:fullcoop}) and case 29 (Figure \ref{fig:fullcoopr}). Hence, these cases are not of much interest. Form all graphs, it is evident that the system in the initial days is not efficient, but as peers start developing resources and social connections, the efficiency increases significantly.   

In competitive strategy (Figure \ref{fig:fullcomp}), for the stationary mobility (Figure \ref{fig:fullcomp} (a)), case 6 and 7 go side by side, whereas, case 8 is more efficient than these two. It means that small-world network with beta = 0.2 is the most efficient setting, and small-world network with a beta value equal to 0.1 is no better than the neighborhood-based regular network. The mobility random walk (Figure \ref{fig:fullcomp} (b)) minimizes the advantage of the small-world network (beta = 0.2), but there is a definite decrease in a number of requests ``nor served" when compared to stationary case. The above is also applicable in case of profile-based mobility (Figure \ref{fig:fullcomp} (c)), however the overall efficiency further increases. It means profile based walk, irrespective of the network type, works the best. It is understandable keeping in view peers moving in the locality and sharing resources in a predefined and repetitive manner, which turns out to be dynamic enough to provide the resources required. Incidentally, this observed behavior supports the mobility model acquired by people in general. 

\begin{figure*}
\centering
\includegraphics[width=.99\textwidth]{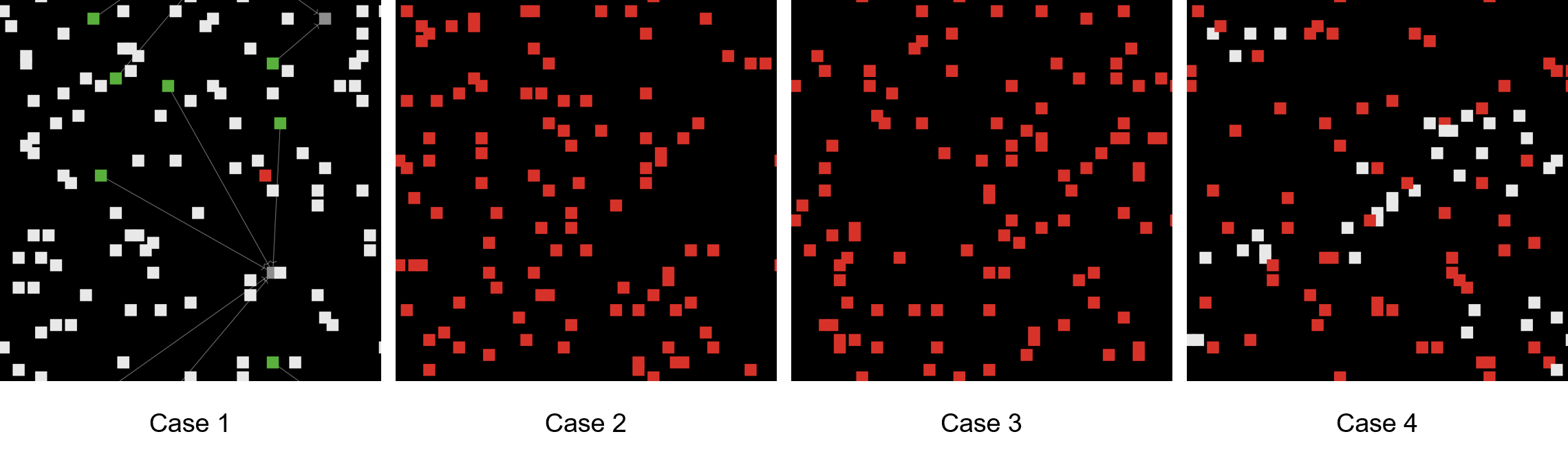}
\caption{Screen shot at the end of the simulation. Competitive mode and 100 agents. Agents in light gray are in offline mode (status 0). Agents in dark gray are in idle mode (status 1). Agents in red are in search mode (status 3). Status 2 is a transit mode where a service is assigned. Agents in green are in request mode (status 4). Status 5 and 6 are transit states again. The arrow originates from the requester and ends at a receiver.}
\label{fig:100compS}
\end{figure*}

\begin{figure*}
\centering
\includegraphics[width=.99\textwidth]{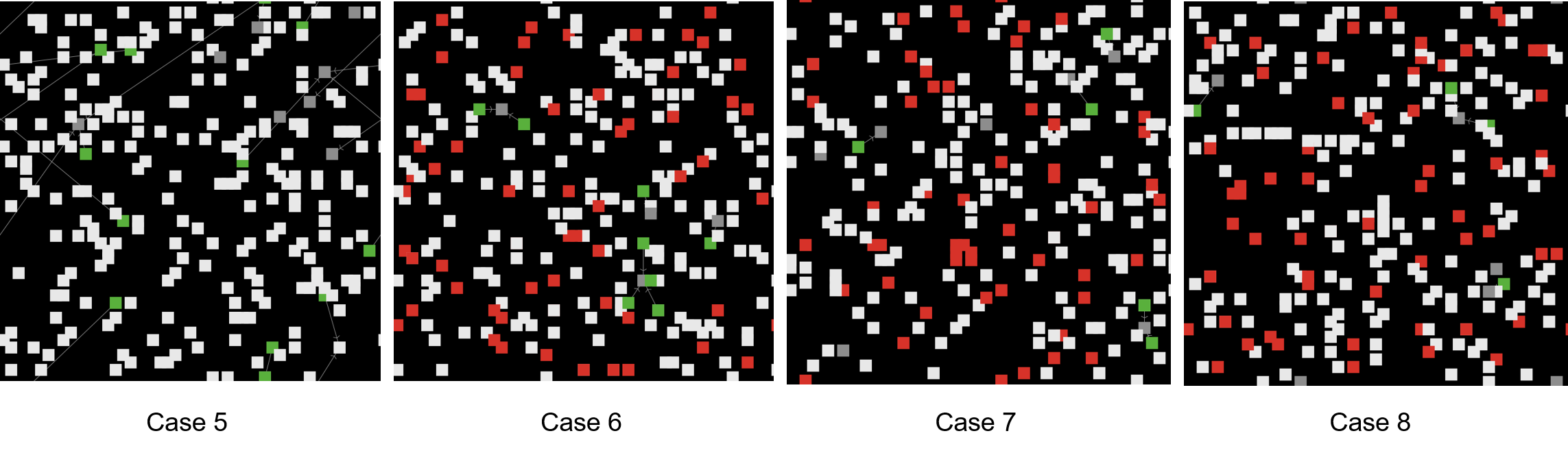}
\caption{Screen shot at the end of the simulation. Competitive mode and 250 agents. Agents in light gray are in offline mode (status 0). Agents in dark gray are in idle mode (status 1). Agents in red are in search mode (status 3). Status 2 is a transit mode where a service is assigned. Agents in green are in request mode (status 4). Status 5 and 6 are transit states again. The arrow originates from the requester and ends at a receiver.}
\label{fig:250compS}
\end{figure*}

\begin{figure*}
\centering
\includegraphics[width=.99\textwidth]{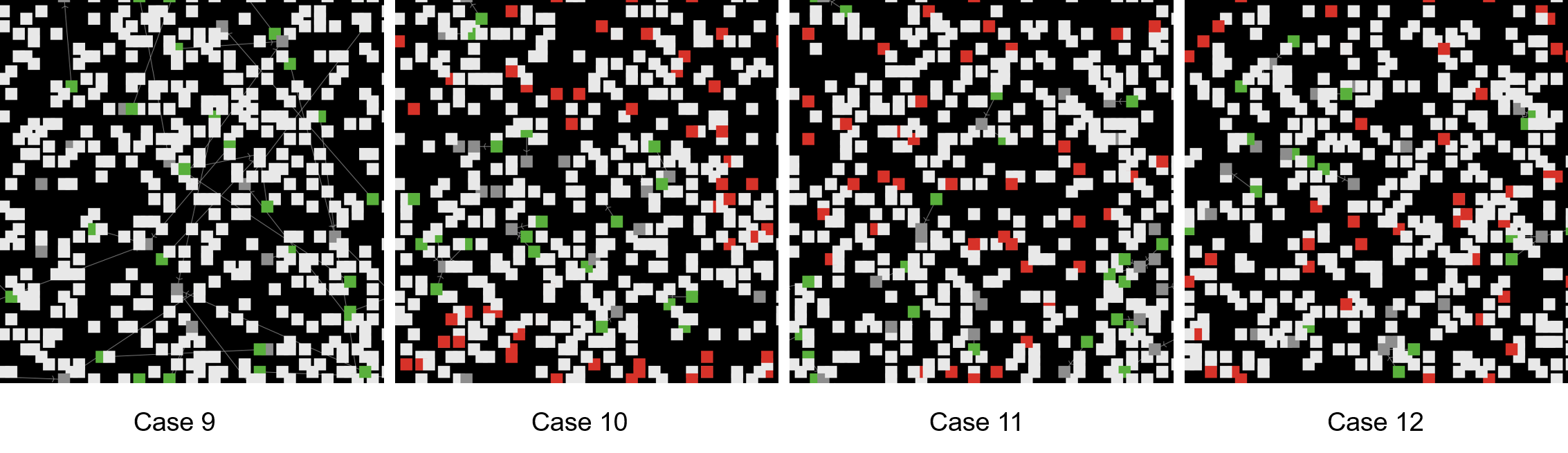}
\caption{Screen shot at the end of the simulation. Competitive mode and 500 agents. Agents in light gray are in offline mode (status 0). Agents in dark gray are in idle mode (status 1). Agents in red are in search mode (status 3). Status 2 is a transit mode where a service is assigned. Agents in green are in request mode (status 4). Status 5 and 6 are transit states again. The arrow originates from the requester and ends at a receiver.}
\label{fig:500compS}
\end{figure*}

\begin{figure} [htb]
\centering
\begin{subfigure}[b]{0.3\textwidth}
\includegraphics[width=\linewidth]{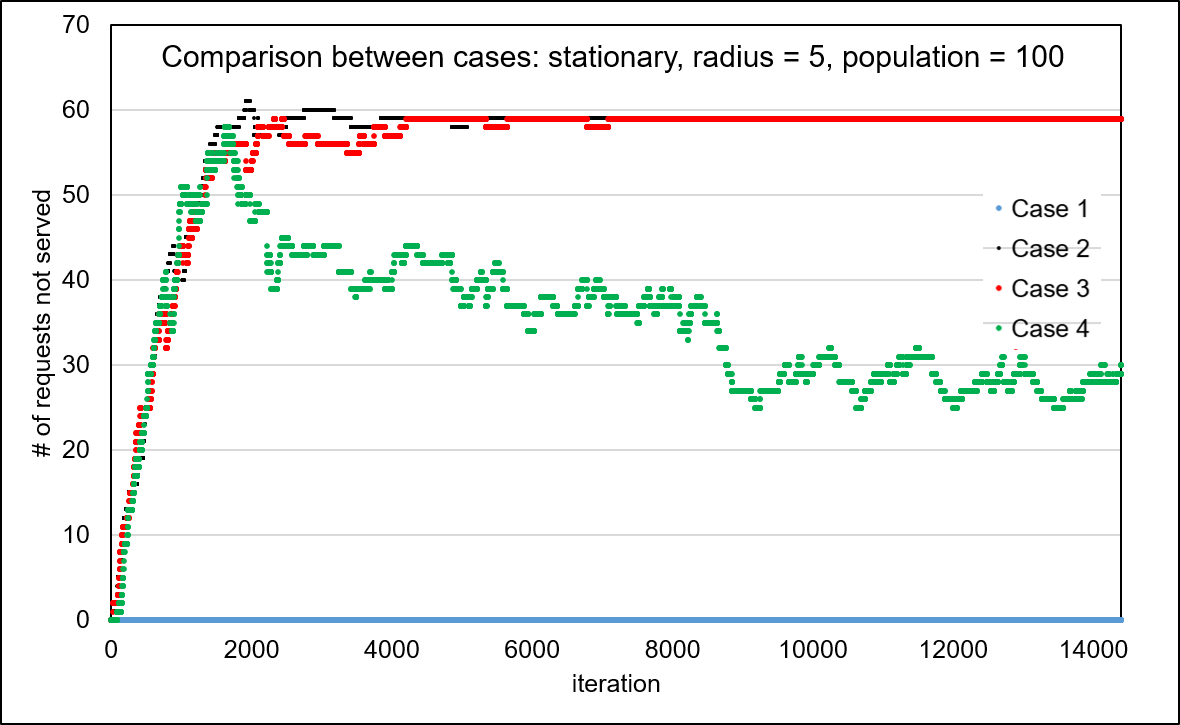}
\label{fig:comp0a}
\end{subfigure}
\hfill
\begin{subfigure}[b]{0.3\textwidth}
\includegraphics[width=\linewidth]{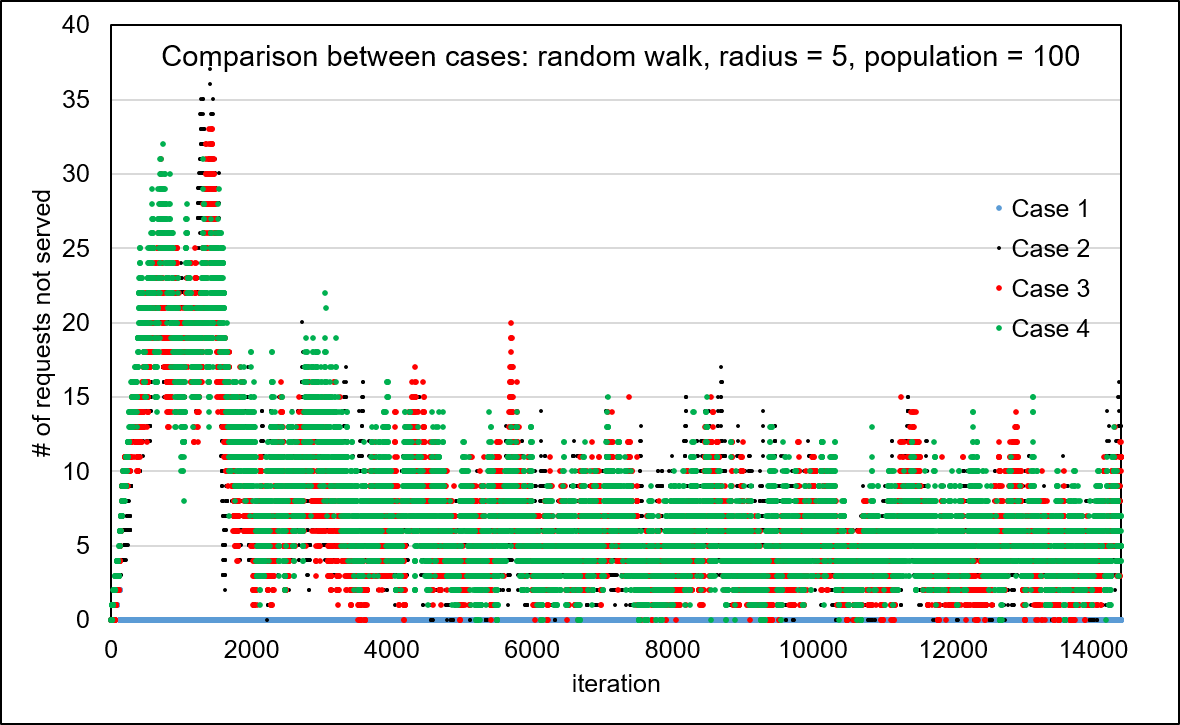}
\label{fig:comp0b}  
\end{subfigure}
\hfill
\begin{subfigure}[b]{0.3\textwidth}
\includegraphics[width=\linewidth]{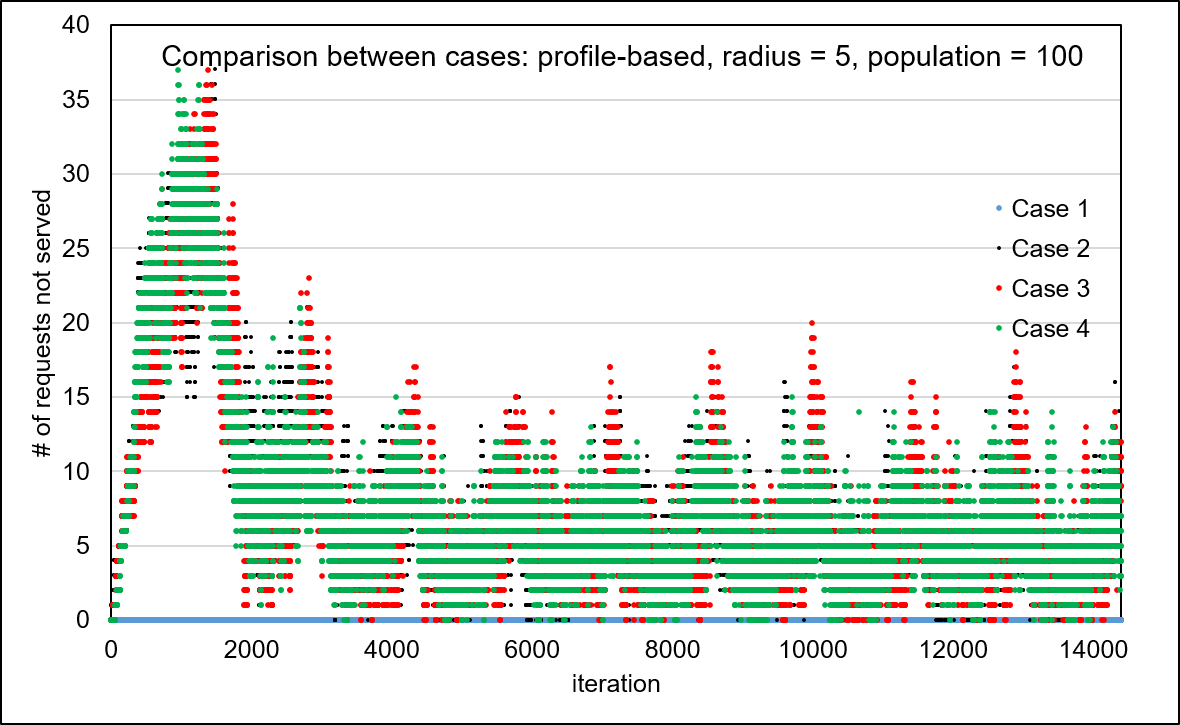}
\label{fig:comp0c}
\end{subfigure}

\caption{Simulation results: Competitive Mode (100 agents)}
\label{fig:fullcomp100}
\end{figure}

\begin{figure}
\centering
\begin{subfigure}[b]{0.3\textwidth}
\includegraphics[width=\linewidth]{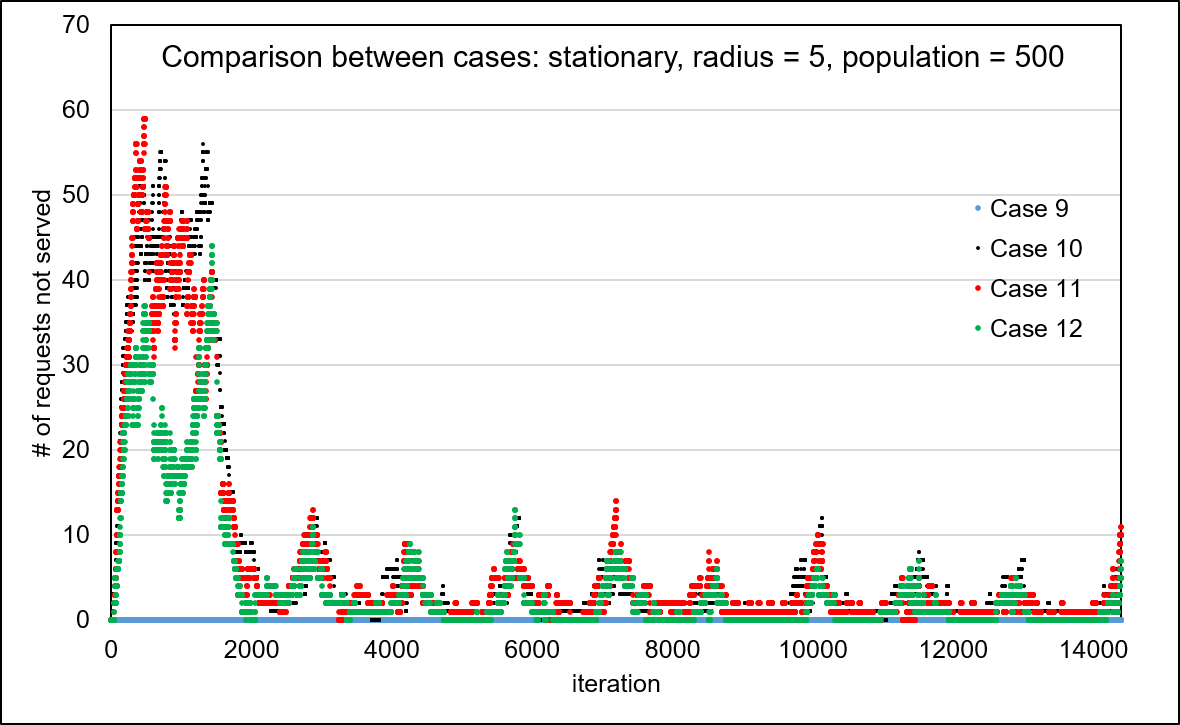}
 \label{fig:comp2a}
\end{subfigure}
\hfill
\begin{subfigure}[b]{0.3\textwidth}
\includegraphics[width=\linewidth]{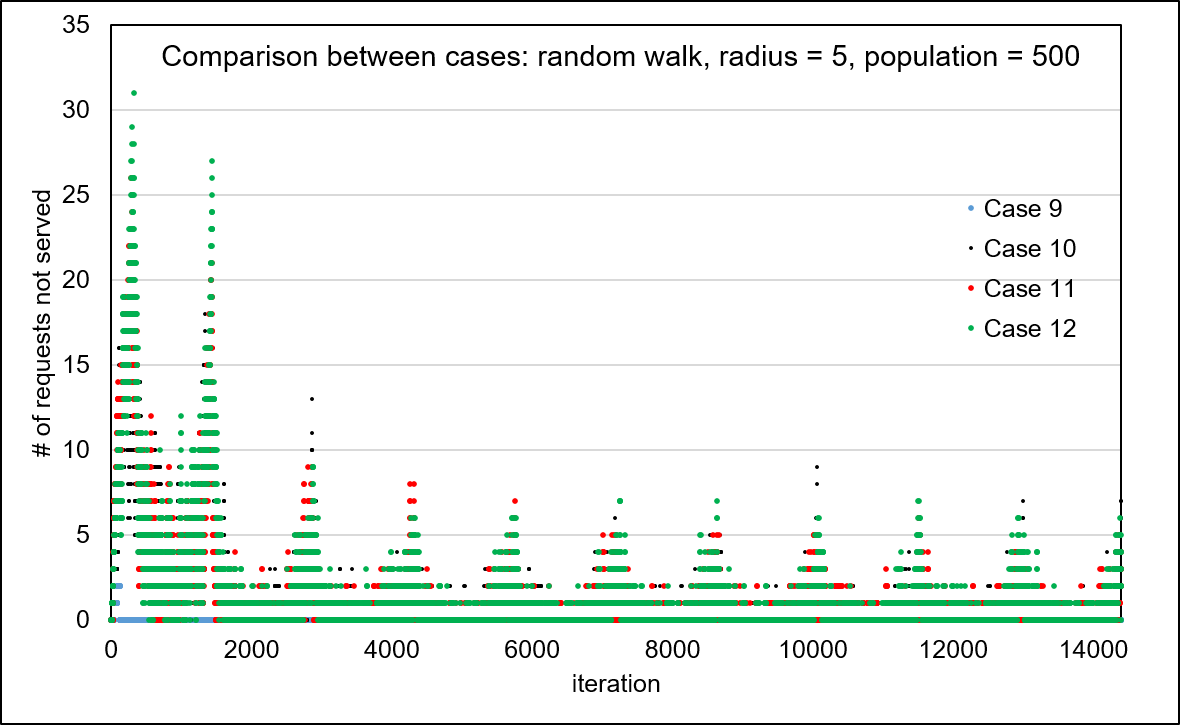}
\label{fig:comp2b}  
\end{subfigure}
 \hfill
\begin{subfigure}[b]{0.3\textwidth}
\includegraphics[width=\linewidth]{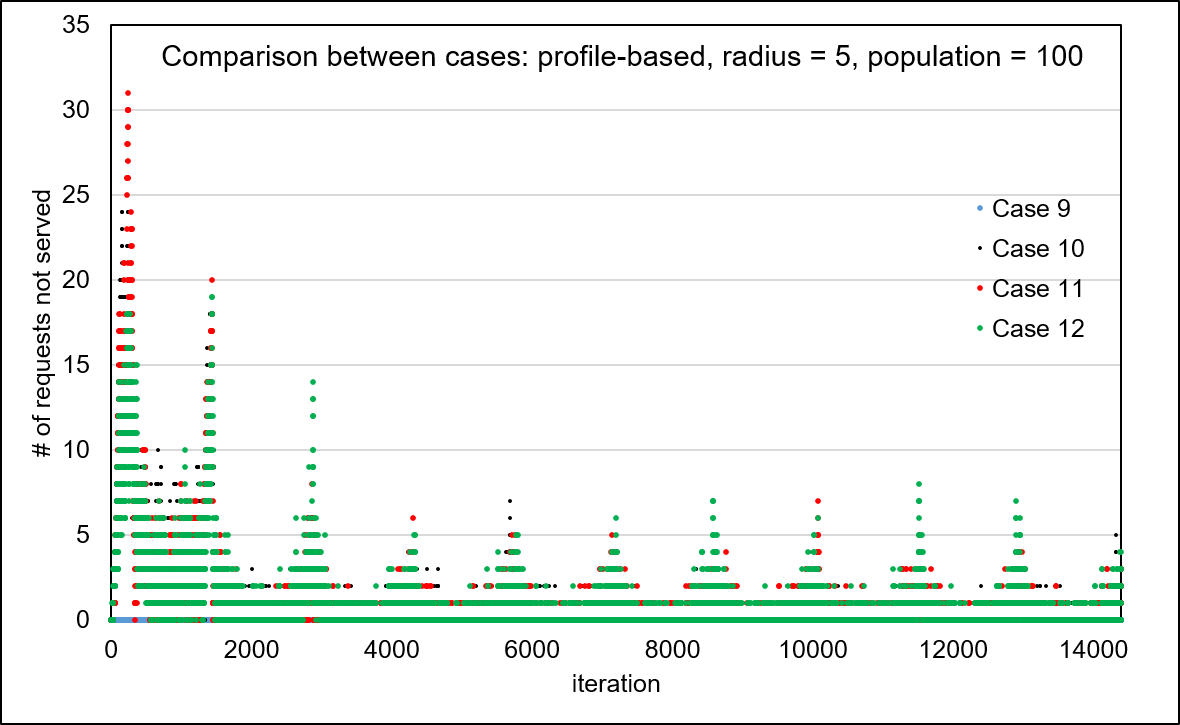}
\label{fig:comp2c}
\end{subfigure}

\caption{Simulation results: Competitive Mode (500 agents)}
\label{fig:fullcomp500}
\end{figure}

In cooperative strategy (Figure \ref{fig:fullcoop}), for the stationary mobility (Figure \ref{fig:fullcoop} (a)), regular network perform worse than the small-world network. But a surprise is that the small-world network with the less beta value performed better than the small-world network with more beta value. However, this can be ignored as the difference is not substantial. When compared to competitive strategy, cooperative strategy is more efficient, particularly in random walk (Figure \ref{fig:fullcoop} (b)) and profile-based mobility (Figure \ref{fig:fullcoop} (c)). 

In cooperative strategy with social restriction (Figure \ref{fig:fullcoopr}), for the stationary mobility (Figure \ref{fig:fullcoopr} (a)), regular network perform worse than the small-world network, but only if beta value is 0.2, not less (0.1). This is an unusual result which needs further investigation. More importantly, this behavior is entirely opposite of cooperative strategy (when comparing regular network with the small-world network of less beta value). When compared to cooperative strategy, this strategy is more efficient, particularly in stationary case, and almost similar in random-walk (Figure \ref{fig:fullcoopr} (b)) and profile-based (Figure \ref{fig:fullcoopr} (c)) mobility. Nevertheless, restricting the neighborhood does not impact efficiency negatively. Rather it remains same or slightly improves. 

Overall, for 250 agents (if the results of the first week (initial perturbation) are ignored), the cooperation in a restricted network performs better than cooperation in the unrestricted network. Competitive strategy cannot compete with any of these. Within one strategy, profile-based mobility performs much better than the stationary and marginally better than the random walk mobility. Figure \ref{fig:250compS} provides screen-shots of sample simulation runs at iteration 14400 in case of 250 agents. 

In case of competitive strategy, as the density of the agents is decreased (see Figure \ref{fig:100compS}), the impact of beta value is much more significant than stationary case (see Figure \ref{fig:fullcomp100}). Whereas, the difference between random walk and profile-based walk is not obvious, although, these two cases are much better than stationary cases. The overall performance decreases the nonavailability of resources. Hence, it is established that the density of nodes in the network should be sufficiently high. 

How high should it be? There is no specific answer provided by the simulation cases that setup, but, there is definitely no drawback of a really high number, for example in case of 500 agents (see Figure \ref{fig:500compS}). In the case of competitive strategy, the high density of agents improves the system in all cases (see Figure \ref{fig:fullcomp500}) when compared to less density. We have observed similar results in the case of cooperative and restricted cooperative strategies. These results are not presented in this paper to reduce redundancy.  

\subsection {Lessons learned from simulation results}

The overall findings are summarized as:

\begin{enumerate}
    \item In a P2P resource sharing scenario, the density of agents (peers) has a significant impact. There is no upper bound of the number of peers. More importantly, it is observed that extremely high density does not deteriorate the performance of the system.
    \item As a whole, cooperation between peers improves the system. In particular, cooperation in a restricted network is never counterproductive; in-fact, it is evidenced to be marginally better than open-ended cooperation. 
    \item Profile-based mobility performs much better than stationary peers and slightly better than random walk mobility.
\end{enumerate}

The above results and overall research outcomes indicate that cooperation in a society of objects of IoT outperforms the competition. In social science, economics and related disciplines, the betterment of the society as a whole are now considered superior to individualistic gains. That's exactly what objects in SIoT are required to do. And designers and practitioners should keep this fact in mind while implementing a solution using this platform.  

Although, digital societies provide the possibility to get access to the whole of the network. But, there is no particular benefit of maintaining such a network in a P2P resource sharing situation. Even, cooperation using restricted network access would suffice. This is another social human feature that we can learn from human society and apply SIoT applications. 

Lastly, peers acquiring more realistic mobility model (the profile-based mobility) works better than stationary and random walk mobility; another social aspect derived from humanistic characteristics.

\section {Conclusions} \label {sec:conc}
A peer-to-peer resource sharing scenario is taken up to analyze the potential of the social Internet of Things. While accepting competitive behavior as a default behavior to model a resource sharing scenario, a model of cooperative behavior is also proposed. It is observed that cooperation (one of the basic social feature) is important for futuristic internet of things constituted by social objects. In addition to cooperation proving to be better than the competition, aspects derived from the social capabilities of peers, that is:
\begin{itemize}
    \item maintaining a restricted network (of friends)
    \item acquiring a realistic mobility model (the profile-based mobility)
\end{itemize}
\noindent benefits the system and turns out to be more efficient than the unrestricted network of peers and unrealistic mobility (in many real-life situations).

\section*{References}

\bibliography{bibio1}


\end{document}